\documentclass[11pt,a4paper]{article}
\usepackage[hyperref]{acl2017}
\usepackage{times}
\usepackage{latexsym}

\usepackage{url}

\usepackage{amsmath,amsfonts,amssymb,bbm,graphicx,tabularx,mathtools}
\usepackage{booktabs}
\aclfinalcopy % Uncomment this line for the final submission
\title{Blind phoneme segmentation with temporal prediction errors}

\author{Paul Michel$^1$\thanks{This work was done when the author was an intern at LSCP / ENS / EHESS / CNRS} \qquad Okko Rasanen$^2$ \qquad Roland Thiolli\`ere$^3$ \qquad Emmanuel Dupoux$^3$ \\
  $^1$Carnegie Mellon University, $^2$Aalto University, $^3$LSCP / ENS / EHESS / CNRS  \\
  {\tt pmichel1@cs.cmu.edu}}

\date{}
\begin{document}
%\ninept
%
\maketitle
\begin{abstract}
Phonemic segmentation of speech is a critical step of speech recognition systems. We propose a novel unsupervised algorithm based on sequence prediction models such as Markov chains and recurrent neural networks. Our approach consists in analyzing the error profile of a model trained to predict speech features frame-by-frame. Specifically, we try to learn the dynamics of speech in the MFCC space and hypothesize boundaries from local maxima in the prediction error. We evaluate our system on the TIMIT dataset, with improvements over similar methods.
\end{abstract}

\section{Introduction}\label{sec:intro}
One of the main difficulty of speech processing as opposed to text processing is the continuous, time-dependent nature of the signal. As a consequence, pre-segmentation of the speech signal into words or sub-words units such as phonemes, syllables or words is an essential first step of a variety of speech recognition tasks.

Segmentation in phonemes is useful for a number of applications (annotation of speech for the purpose of phonetic analysis, computation of speech rate, keyword spotting, etc), and can be done in two ways. Supervised methods are based on an existing phoneme or word recognition system, which is used to decode the incoming speech into phonemes. Phonemes boundaries can then be extracted as a by-product of the alignment of the phoneme models with the speech. Unsupervised methods (also called blind segmentation) consist in finding phonemes boundaries using the acoustic signals only. Supervised methods depend on the training of acoustic and language  models, which requires access to large amounts of linguistic resources (annotated speech, phonetic dictionary, text). Unsupervised methods do not require these resources and are therefore appropriate for so-called under-resourced languages, such as endangered languages, or languages without consistent orthographies. 

We propose a blind phoneme segmentation method based on short term statistical properties of the speech signal. We designate peaks in the error curve of a model trained to predict speech frame by frame as potential boundaries. Three different models are tested. The first is an approximated Markov model of the transition probabilities between categorical speech features. We then replace it by a recurrent neural network operating on the same categorical features. Finally, a recurrent neural network is directly trained to predict the raw speech features. This last model is especially interesting in that it couples our statistical approach with more common spectral transition based methods (\citet{dusan2006relation} for instance). 

We first describe the various models used and the pre- and post-processing procedures, before presenting and discussing our results in the light of previous work.

\section{Related work}\label{sec:related}

Most previous work on blind phoneme segmentation \cite{esposito2005text, estevan2007finding, almpanidis2008phonemic, rasanen2011blind,khanagha2014phonetic, hoang2015blind} has focused on the analysis of the rate of change in the spectral domain. The idea is to design robust acoustic features that are supposed to remain stable within a phoneme, and change when transitioning from one phoneme to the next. The algorithm then define a measure of change, which is then used to detect phoneme boundaries.

Apart from this line of research, three main approaches have been explored. The first idea is to use short term statistical dependencies. In  \citet{rasanen2014basic}, the idea was to first discretize the signal using a clustering algorithm and then compute discrete sequence statistics, over which a threshold can be defined. This is the idea that we follow in the current paper. The second approach is to use dynamic programming methods inspired by text segmentation \cite{wilber1988concave}, in order to derive optimal segmentation  \cite{qiao2008unsupervised}. In this line of research, however, the number of segments is assumed to be known in advance, so this cannot count as blind segmentation. The third approach consists in jointly segmenting and learning the acoustic models for phonemes \cite{kamper2015fully, glass2003probabilistic,siu2013}. These models are much more computationally involved than the other methods. Interestingly they all use a simpler, blind segmentation as an initialization phase. Therefore, improving on pure blind segmentation could be useful for joint models as well.

The principal source of inspiration for our work comes from previous work by \citet{elman1990finding} and \citet{christiansen1998learning} published in the 90s. In the former, the author uses recurrent neural networks to train character-based language models on text and notices that "The error provides a good clue as to what the recurring sequences in the input are, and these correlate highly with words." \cite{elman1990finding}.
More precisely, the error tends to be higher at the beginning of new words than in the middle.
In the latter, the author uses Elman recurrent neural networks to predict boundaries between words given the character sequence and phonological cues.

Our work uses the same idea, using prediction error as a cue for segmentation, but with two important changes: we apply it to speech instead of text, and we use it to segment in terms of phoneme units instead of word units.

\section{System}

\subsection{Pre-processing}
We used two kinds of speech features : 13 dimensional MFCCs \cite{davis1980comparison} (with 12 mel-cepstrum coefficients and 1 energy coefficient) and categorical one-hot vectors derived from MFCCs inspired by \citet{rasanen2014basic}.

\begin{figure}[!ht]
    \centering
    \includegraphics[width=\columnwidth]{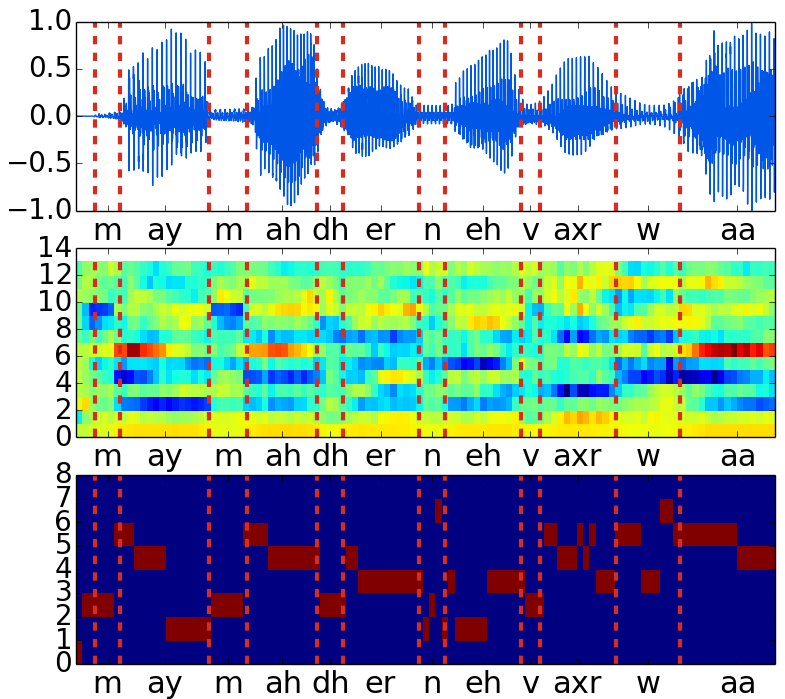}
    \caption{Visual representation of the various features on 100 frames from the TIMIT corpus. From top to bottom are the waveform, the 13-dimensional MFCCs and the 8-dimensional one hot encoded categorical features.}
    \label{fig:feats}
\end{figure}

The latter are computed according to \citet{rasanen2014basic} : K-means clustering\footnote{In particular, we use the K-means++ \cite{arthur2007k} algorithm, and pick the best outcome of 10 random initializations} is performed on a random subset of the MFCCs (10,000 frames were selected at random), with a target number of clusters of 8, then each MFCC is identified to the closest centroid. Each frame is then represented by a cluster number $c\in \{1,\ldots,8\}$, or alternatively by the corresponding one-hot vector of dimension $8$. These hyper-parameters were chosen according to \citet{rasanen2014basic}.

Figure \ref{fig:feats} allows for a visual comparison of the three signals (waveform, MFCC, categorical).

The entire dataset is split between a training and a testing subset. A randomly selected subset of the training part is used as validation data to prevent overfitting.

\subsection{Training phase}
A frame-by-frame prediction model is then learned on the training set. The three different models used are described below : 

\paragraph{Pseudo-markov model}

When trying to predict the frame $x_t$ given the previous frames $x_0^{t-1}\coloneqq x_{t-1},\ldots,x_{0}$, a simplifying assumption is to model the transition probabilities with a Markov chain of higher order $K$, i.e. $p(x_t|x_{0}^{t-1})=p(x_t\mid x_{t-K}^{t-1})$. Provided each frame is part of a finite alphabet, a finite (albeit exponential in $K$) number of transition probabilities must be learned.

However, as the order rises, the ratio between the size of the data and the number of transition probability being learned makes the exact calculation more difficult and less relevant.

In order to circumvent this issue, we approximate the $K$-order Markov chain with the mean of 1-order markov chain of the lag-transition probabilities $p(x_t\vert x_{t-i})$ for $1\leqslant i \leqslant K$, so that 
\begin{equation}p(x_t|x_0^{t-1})=\frac 1 K \sum_{i=1}^K p(x_t\vert x_{t-i})
\end{equation} with $p(x_t\vert x_{t-i})=\frac {f(x_t,x_{t-i})}{f(x_{t-i})}$.

In practice, we chose $K=6$, thus ensuring that the markov model's attention is of the same order of magnitude than the length of a phoneme.

Compared to \citet{rasanen2014basic}, this model only uses information from previous frames and as such is completely online.

\paragraph{Recurrent neural network on categorical features}

Alternatively to Markov chains, the transition probability $p(x_t\vert x_0^{t-1})$ can be modeled by a recurrent neural network (RNN). RNN can theoretically model indefinite order temporal dependencies, hence their advantage over Markov chains for long sequence modeling.

Given a set of examples $\{(x_t,(x_0^{t-1}))\mid t\in \{0,\ldots,t_{max}\}\}$, the networks parameters are learned so that the error $E(x_t,\text{RNN}(x_0^{t-1}))$ is minimized using back propagation through time \cite{werbos1990backpropagation} and stochastic gradient descent or a variant thereof (we have found RMSProp \cite{tieleman2012lecture} to give the best results).

In our case, the network itself consists of two LSTM layers \cite{hochreiter1997long} stacked on one another followed by a linear layer and a softmax. The input and output units have both dimension 8, whereas all other layers have the same hidden dimension 40. Dropout \cite{srivastava2014dropout} with probability 0.2 was used after each LSTM layer to prevent overfitting.

A pitfall of this method is the tendency of the network to predict the last frame it is fed. This is due to the fact that the sequences of categorical features extracted from speech contain a lot of constant sub-sequences length $\geqslant 2$.

As a consequence, around 80\% of the data fed to the network consists of sub-sequences
where $x_t = x_{t-1}$ . Despite the fact that phone boundaries are somewhat correlated with changes of categories (around 65\% of the time), this leads the network to a local minimum where it only tries to predict the same characters.

To mitigate this effect, examples where $x_t=x_{t-1}$ were removed with probability $0.8$, so that the number of transitions was slightly skewed towards category transitions. The model still passed over all frames during training but the error was back-propagated for only 46\% of them. This change lead to substantial improvement.

\paragraph{Recurrent neural network on raw MFCCs}

The recurrent neural network model can be adapted to raw speech features simply by changing the loss function from categorical cross-entropy to mean squared error, which is the direct translation from a categorical distribution to a Gaussian density ($2\Vert \textbf{x}-\textbf{y}\Vert_2^2+d$ is the Kullback-Leibler divergence of two $d$-dimensional normal distributions centered in $\textbf{x}$ and $\textbf{y}$ with the same scalar covariance matrix).

We used the same architecture than in the categorical case, simply removing the softmax layer and decreasing the hidden dimension size to 20. In this case, no selection of the samples is needed since the sequences vary continuously.

\subsection{Test phase}

Each model is run on the test set and the prediction error is calculated at each time step according to the formula :

\begin{equation}
\begin{split}
    \text{E}_{\text{markov}}(t)&=-\log\left(\sum_{i=1}^\mathrm{K}p(x_t\vert x_{t-i})\right)\\
    \text{E}_{\text{\tiny RNN-cat}}(t)&=-\sum_{i=1}^d\mathbbm 1_{x_t=i}\log(\text{RNN}(x_0^{t-1}))\\
    \text{E}_{\text{\tiny RNN-MFCC}}(t)&=\frac 1 d \left\Vert \textbf{x}_t-\text{RNN}(\textbf{x}_0^{t-1})\right\Vert _2 ^ 2\\
\end{split}
\end{equation}

In each case this corresponds, up to a scaling factor constant across the dataset, to the Kullback-Leibler divergence between the predicted and actual probability distribution for $x_t$ in the feature space.

Since all three systems predict probabilities conditioned by the preceding frames, they cannot be expected to give meaningful results for the first frames of each utterance. To be consistent, the first 7 frames (70 ms) of the error signal for each utterance were set to 0.

A peak detection procedure is then applied to the resulting error. As we are looking for sudden bursts in the prediction error, a local maximum is labeled as a potential boundary if and only if the difference between its value and the one of the previous minimum is superior to a certain threshold $\delta$.

\section{Experiments}

\subsection{Dataset}
\begin{table}[t]
    \centering
    \begin{tabular}{|c|c|c|c|c|}
        \hline
        Algorithm & P & R & F & R-val\\
        \hline
        Periodic & 57.5 & 91.0 & 70.5 & 46.9\\
        \hline
        Rasanen \shortcite{rasanen2014basic}& 68.4 & 70.6 & 69.5 & 73.7 \\
        \hline
        Markov & 70.7 & 77.3 & 73.9 & 76.4 \\
        \hline
        RNN (Cat.) & 68.7 & 77.1 & 72.7 & 74.6 \\
        \hline
        RNN (Cont.) & 70.3 & 72.4 & 71.3 & 75.3\\
        \hline
    \end{tabular}
    \caption{Final results (in\%) evaluated with cropped tolerance windows}
    \label{tab:res}
\end{table}

We evaluated our methods on the TIMIT dataset \citet{TIMIT}. The TIMIT dataset consists of 6300 utterances ($\sim$ 5.4 hours) from 630 speakers spanning 8 dialects of the English language. The corpus was divided into a training and test set according to the standard split. The training set contains 4620 utterances (172,460 boundaries) and the test set 1680 (65,825 boundaries).

\subsection{Evaluation}

The performance evaluation of our system is based on precision ($P$), recall ($R$) and $F$-score, defined as the harmonic mean of precision and recall. A drawback of this metric is that high recall, low precision results, such as the ones produces by hypothesizing a boundary every 5 ms (P : 58\%, R : 91\%) yield high $F$-score (70\%).

Other metrics have been designed to tackle this issue. One such example is the R-value \cite{rasanenrval2009} :
\begin{equation}
\text{R-val}=1-\frac{\sqrt{(1-\text{R})^2+\text{OS}^2}+\vert\frac{\text{R} + 1-\text{OS}}{\sqrt{2}}\vert}{2}
\end{equation}

Where $\text{O}S=\frac {\text{R}} {\text{P}} - 1$ is the over-segmentation measure. The R value represents how close the segmentation is from the ideal 0 OS, 1 R point and the P=1 line in the R, OS space. Further details can be found in \citet{rasanenrval2009}.

Determining whether gold boundary is detected or not is a crucial part of the evaluation procedure. On our test set for instance, which contains 65,825 gold boundaries partitioned into 1,680 files, adding or removing one correctly detected boundary per utterance leads to a change of $\pm$ 2.5\% in precision. This means that minor changes in the evaluation process (such as removing the trailing silence parts of each file, removing the opening and closing boundary) yield non-trivial variations in the end result.

A common condition for a gold boundary to be considered as 'correctly detected' is to have a proposed boundary within a 20 ms distance on either side. Without any other specification, this means that a proposed boundary may be matched to several gold boundaries, provided these are within 40 ms from each other, leading to an increase of up to 4\% F-score in some of our results (74\%---78\%). Unfortunately this point is seldom detailed in the literature.

We decided to use the procedure described in \citet{rasanenrval2009} to match gold boundaries and hypothesized boundaries : overlapping tolerance windows are cropped in the middle of the two boundaries.

\subsection{Results}

\begin{table}[!t]
    \centering
    \begin{tabular}{|c|c|c|c|c|}
        \hline
        Algorithm & P & R & F & R-val\\
        \hline
        Periodic & 62.2 & 98.3 & 76.2 & 49.8\\
        \hline
        Rasanen \shortcite{rasanen2014basic}& 74.0 & 70.0 & 73.0 & 76.0\\
        \hline
        Markov & 74.8 & 81.9 & 78.2 & 80.1\\
        \hline
        RNN (Cat.) & 72.5 & 81.4 & 76.7 & 78.0\\
        \hline
        RNN (Cont.) & 77.6 & 72.7 & 75.0 & 78.6\\
        \hline
    \end{tabular}
    \caption{Final results (in\%) evaluated with overlapping tolerance windows. The scores reported for \textit{Rasanen (2014)} are the paper results.}
    \label{tab:res_over}
\end{table}

The current state of the art in blind phoneme segmentation on the TIMIT corpus is provided by \citet{hoang2015blind}. It evaluates to 78.16\% F-score and 81.11 R-value on the training part of the dataset, using an evaluation method similar to our own.

In Tables \ref{tab:res} and \ref{tab:res_over} we compare our best results to the previous statistical approach evoked in \citet{rasanen2014basic} and the naive periodic boundaries segmentation (one boundary each 5 ms). Since \citet{rasanen2014basic} used an evaluation method allowing for tolerance windows to overlap, we provide our results with both evaluation methods (full windows and cropped windows) for the sake of consistency.

Another main difference with \citet{rasanen2014basic} is that its results are given on the core test set of TIMIT, whereas our results are given on the full test set.

\begin{figure}[!ht]
    \centering
    \includegraphics[width=\columnwidth]{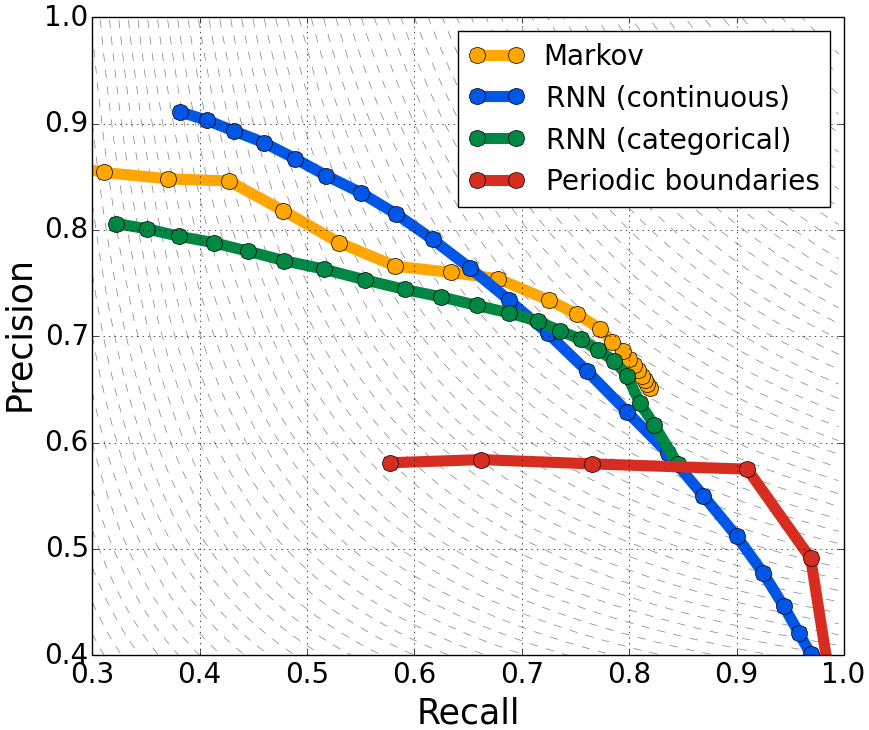}
    \caption{Precision/recall curves for our various models when varying the peak detection threshold $\delta$}
    \label{fig:recap}
\end{figure}

Figure \ref{fig:recap} provides an overview of the precision/recall scores when varying the peak detection threshold (and, in case of periodic boundaries, the period). This gives some insight about the actual behavior of the various algorithms, especially in the high precision, low recall region where the RNN on actual MFCCs seems to outperform the methods based on discrete features.

\begin{figure}[!ht]
    \centering
    \includegraphics[width=\columnwidth]{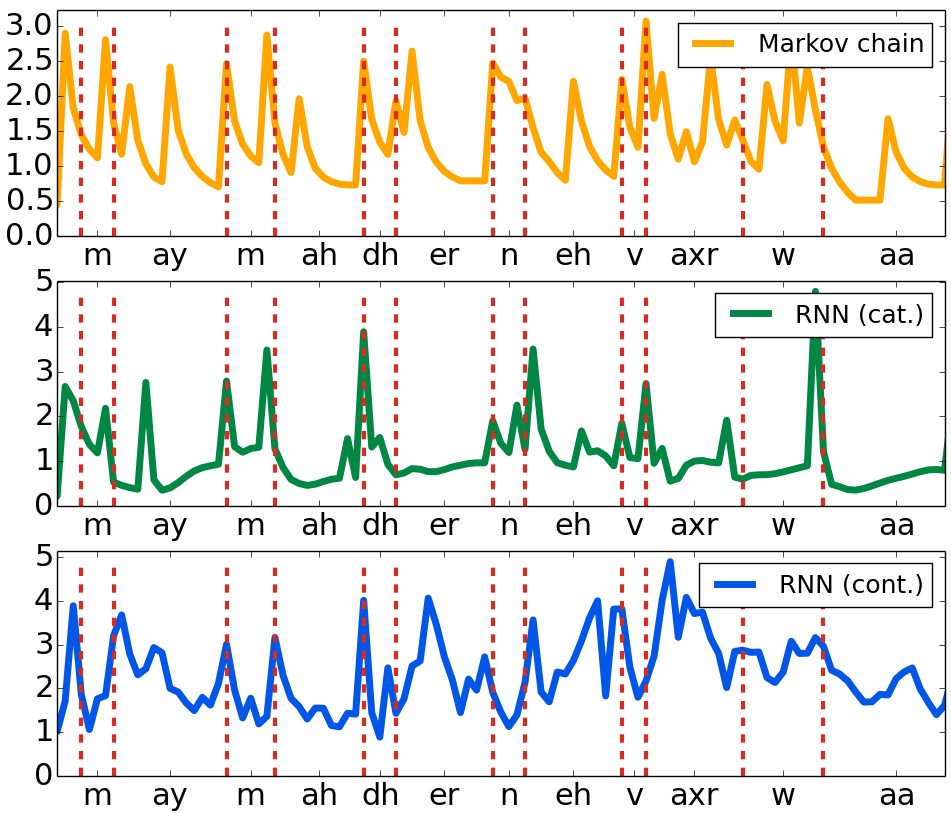}
    \caption{Comparison of error signals (gold boundaries are indicated in red)}
    \label{fig:qual}
\end{figure}

We provide Figure \ref{fig:qual} as a qualitative assessment of the error profiles of all three algorithms on one specific utterance. Notably, the error profile of the markov model contains distinct isolated peaks of similar height. As expected, the error curve is much more noisy in the case of the RNN on MFCCs, due to the greater variability in the feature space.

\section{Discussion}

In terms of optimal F-score and R values, the simple Markov model outperformed the previously published paper using short term sequential statistics \cite{rasanen2014basic}, as well as the recurrent neural networks. However, these optimal values may mask the differential behavior of these algorithms in different sections of the precision/recall curve. In particular, it is interesting to notice that the neural network based model trained on the raw MFCCs gave very good results in the low recall, high precision domain. Indeed, the precision can reach 90\% with a recall of 40\%. Such a regime could be useful, for instance, if blind phoneme segmentation is used to help with word segmentation. 

The reason of the higher precision of neural networks may be that it combines the sensitivity of this model to sequential statistical regularities of the signal, but also to the spectral variations, i.e. the error is also correlated to the spectral changes, meaning that some peaks are associated with a high error because the euclidean distance $\Vert x_{t+1}-x_{t}\Vert_2$ itself is big. This is why the height difference is much more significant in this case.

Although we only reported the best results, we also tested our model on two other neural network architectures : a single vanilla RNN and a single LSTM cell. Both architecture did not yield significantly different results ($\sim$ 1---2\% F-score, mainly dropping precision). Similarly, different hidden dimension were tested. In the extreme cases (very low - 8 - or high - 128 - dimension), the output signal proved too noisy to be of any significance, yielding results comparable to naive periodic segmentation.

It is worth mentioning that our approach doesn't make any language specific assumption, and as such similar results are to be expected on other languages. We leave the confirmation of this assumption to future work.

\section{Conclusions}
\label{sec:conclusions}

We have presented a lightweight blind phoneme segmentation method predicting boundaries at peaks of the prediction loss of transition probabilities models. The different models we tested produced satisfying results while remaining computationally tractable, requiring only one pass over the data at test time.

Our recurrent neural network trained on speech features in particular hints at a way of combining both the statistical and spectral information into a single model.

On a machine learning point of view, we highlighted the use that can be made of side channel information (in this case the test error) in order to extract structure from raw data in an unsupervised setting.

Future work may involve exploring different RNN models, assessing the stability of these methods on simpler features such as raw spectrograms or waveforms, or exploring the representation of each frame in the hidden layers of the networks.

\section{Acknowledgements}

The authors would like to thank the anonymous reviewers for their insightful and constructive comments which helped shape the final version of this paper.

This project is supported by the European Research Council (ERC-2011-AdG-295810 BOOTPHON), the Agence Nationale pour la Recherche (ANR-10-LABX-0087 IEC, ANR-10-IDEX-0001-02 PSL*), the Fondation de France, the Ecole de Neurosciences de Paris, the Region Ile de France (DIM cerveau et pens\'{e}e), and an AWS in Education Research Grant award.

\bibliographystyle{acl_natbib}
\bibliography{refs}

\end{document}